\documentclass[11pt,a4paper]{article}
\usepackage[hyperref]{acl2020}
\usepackage{times}
\usepackage{latexsym}

\usepackage{amsmath, amssymb, amsthm}
\usepackage{xspace}
\usepackage{booktabs}
\usepackage{graphicx}
\usepackage{multirow}
\usepackage{xcolor}
\usepackage{subcaption}
\usepackage{tikz}

\newcommand{\bpeEmb}{byte-pair encoding\xspace}
\newcommand{\flairEmb}{FLAIR\xspace}
\newcommand{\metaEmb}{meta-embeddings\xspace}

\newcommand{\dPerp}{$d_{P}$\xspace}
\newcommand{\dPerpFlair}{$d_{P}^{*}$\xspace}
\newcommand{\dVocab}{$d_{V}$\xspace}
\newcommand{\dVocabTrain}{$d_{V}^{*}$\xspace}
\newcommand{\dVoting}{$d_{MV}$\xspace}

\newcommand{\fs}{$^*$\xspace}

\newcommand{\ssBase}{$^\dagger$\xspace}

\newcommand{\bestPred}[1]{\setlength{\fboxrule}{0.9pt} \fcolorbox{white}{gray!30}{#1}}
\newcommand{\bestReal}[1]{\setlength{\fboxrule}{0.9pt} \fbox{#1}}
\newcommand{\bestBoth}[1]{\setlength{\fboxrule}{0.9pt} \fcolorbox{black}{gray!30}{#1}}

\aclfinalcopy
\def\aclpaperid{34}

\title{On the Choice of Auxiliary Languages for Improved Sequence Tagging}

\author{Lukas Lange$^{1,2,3}$ \\
	\And
	Heike Adel$^1$ \\
	$^1$ Bosch Center for Artificial Intelligence, Renningen, Germany\\
	$^2$ Spoken Language Systems (LSV), Saarland University, Saarbr\"{u}cken, Germany\\
	$^3$ Saarbr\"{u}cken Graduate School of Computer Science, Saarbr\"{u}cken, Germany\\
	\texttt{\{Lukas.Lange,Heike.Adel,Jannik.Stroetgen\}@de.bosch.com}
	\And
	Jannik Str\"{o}tgen$^1$ \\
	\\}

\date{}

\begin{document}
\maketitle
\begin{abstract}
	Recent work showed that embeddings from related languages can improve the performance of sequence tagging, even for monolingual models.
	In this analysis paper, we investigate 
	whether the best auxiliary language can be predicted based on language distances and show that the most related language is not always the best auxiliary language.
	Further, we show that attention-based \metaEmb can effectively combine pre-trained embeddings from different languages for sequence tagging
	and set new state-of-the-art results for part-of-speech tagging in five languages. 
\end{abstract}

\section{Introduction}\label{sec:introduction}
State-of-the-art methods for sequence tagging tasks, such as named entity recognition (NER) and part-of-speech (POS) tagging, exploit embeddings as input representation. 
Recent work suggested to include embeddings trained on related languages
as auxiliary embeddings to improve model performance: 
Catalan and Portuguese embeddings, for instance, help
NER models on Spanish-English code-switching data \cite{emb/Winata19}. 
In this paper, we analyze whether auxiliary embeddings should be chosen from related languages, or if embeddings from more distant languages could also help. 

For this, we revisit current language distance measures~\cite{language/Gamallo17} and adapt them to the embeddings and training data used in our experiments. 
We investigate the question, whether we can predict the best auxiliary language based on those language distance measures.
Our results suggest that no strong correlation exists between language distance and performance and that even less related languages can be a good choice as auxiliary languages.

In our experiments, we explore both available monolingual and multilingual pre-trained \bpeEmb~\cite{bpemb/heinzerling18} and \flairEmb embeddings~\cite{flair/Akbik18}. For combining monolingual subword embeddings from different languages, we investigate two different methods: the concatenation of embeddings and the use of attention-based \metaEmb~\cite{kiela-etal-2018-dynamic, emb/Winata19}.

We perform experiments 
on CoNLL and universal dependency
datasets for NER and POS tagging in five languages and show that \metaEmb are a promising alternative to the concatenation of additional auxiliary embeddings as they learn to decide on the auxiliary languages in an unsupervised way.
Moreover, the inclusion of more
languages is often beneficial and \metaEmb can be effectively used to leverage a larger number of embeddings and achieve new state-of-the-art performance on all five POS tagging tasks. 
Finally, we propose guidelines to decide which auxiliary languages one should use in which setting. 
 
\begin{figure*}[t]
	\centering
	\begin{subfigure}[t]{0.3\textwidth}
		\centering
		\includegraphics[width=1.0\textwidth]{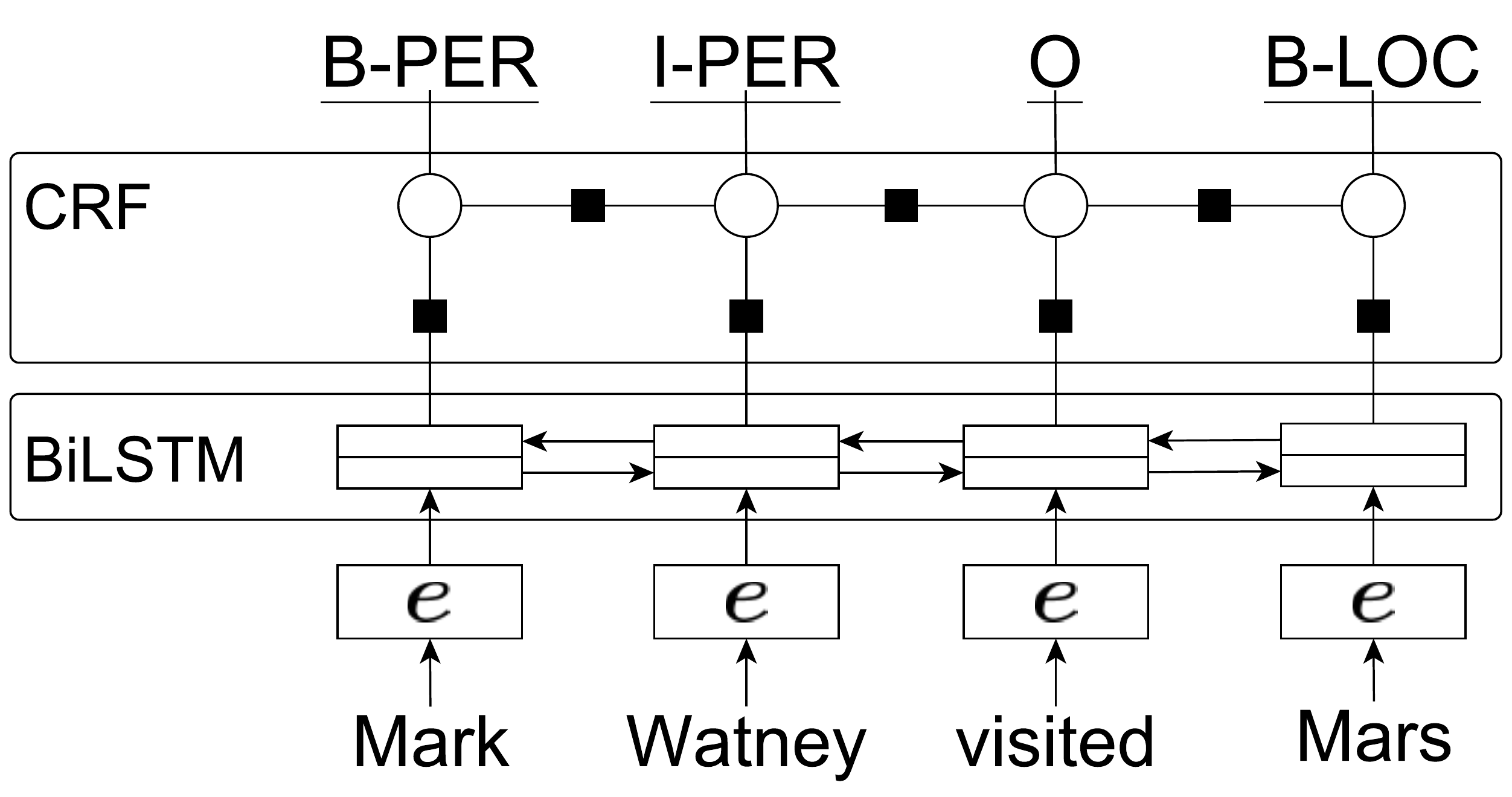}
		\caption{BiLSTM-CRF.}
		\label{fig:model_base}
	\end{subfigure}
	\enskip
	\begin{subfigure}[t]{0.3\textwidth}
		\centering
		\includegraphics[width=1.0\textwidth]{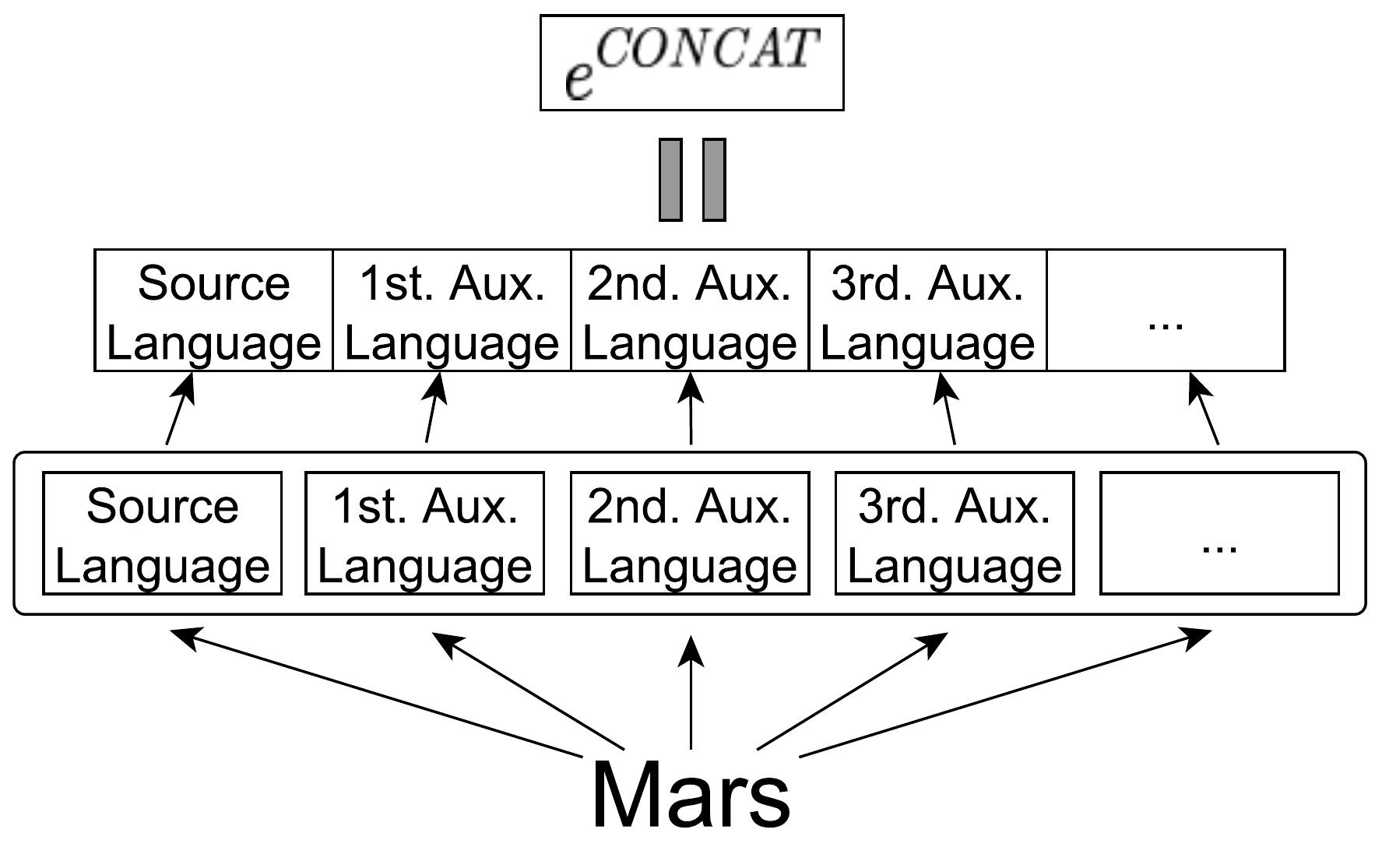}
		\caption{Concatenation.}
		\label{fig:model_concat}
	\end{subfigure}
	\enskip
	\begin{subfigure}[t]{0.3\textwidth}
		\centering
		\includegraphics[width=1.0\textwidth]{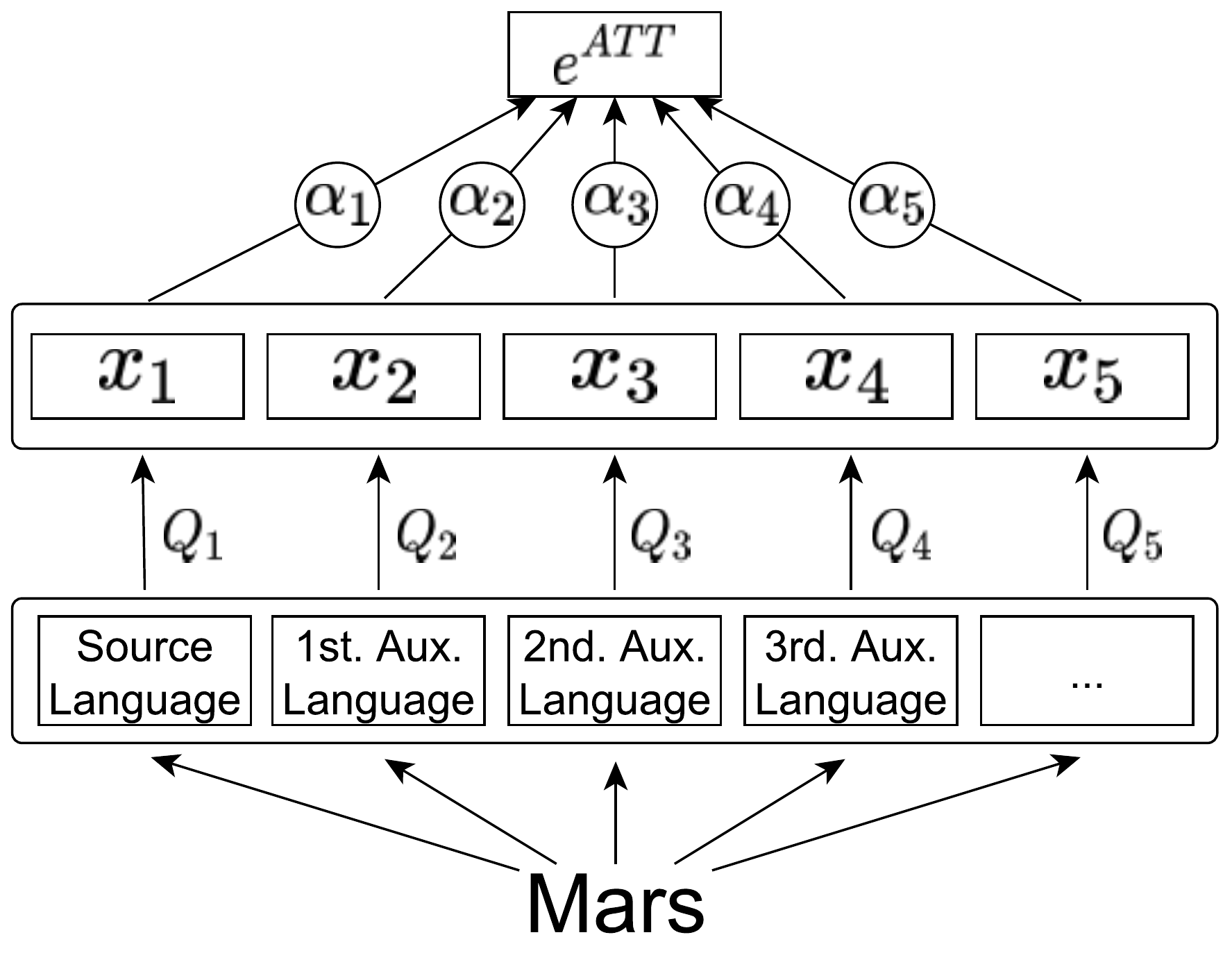}
		\caption{Meta Embedding.}
		\label{fig:model_meta}
	\end{subfigure}
	\caption{Overview of our model architecture (left). The embedding combination $e$ can be either computed using the concatenation $e^{CONCAT}$ (middle) or the meta embedding method $e^{ATT}$ (right).}
	\label{fig:models}
\end{figure*}

\section{Related Work}\label{sec:related}

\paragraph{Combination of Embeddings.}
\label{sec:relatedWorkEmbeddings}
Previous work has seen performance gains by combining, e.g., various types of word embeddings \cite{tsuboi-2014-neural} or variants of the same type of embeddings trained on different corpora \cite{luo2014}.
For the combination, alternative solutions have been proposed, 
such as 
different input channels
\cite{zhang-etal-2016-mgnc}, 
concatenation
\cite{yin-schutze-2016-learning}, averaging of embeddings
\cite{coates-bollegala-2018-frustratingly}, 
and attention
\cite{kiela-etal-2018-dynamic}. 
In this paper, we compare the inclusion of auxiliary languages via concatenation to the dynamic combination with attention.

\paragraph{Auxiliary Languages.}
\newcite{emb/Winata19} proposed to include embeddings from closely-related languages to improve NER performance in code-switching settings, i.e., it was shown that Catalan and Portuguese embeddings help for Spanish-English NER. 
In a later work, it was shown that also more distant languages can be beneficial~\cite{winata-etal-2019-hierarchical}, but only tests in the special setting of code-switching NER were performed and no connection between the relatedness of languages and the performance increase was analyzed. 
In contrast, our work shows that the inclusion of auxiliary languages increases performance in monolingual settings as well and we analyze whether language distance measures can be used to select the best auxiliary language in advance. 

\paragraph{Language Distance Measures.}
\newcite{language/Gamallo17} proposed to measure distances between languages by using the perplexity of language models trained on one language and applied to another language. 
\newcite{campos2019} 
used a similar method to retrace changes in multilingual diachronic corpora over time. 
Another popular measure of similarity is based on vocabulary overlap, assuming that similar languages share a large portion of their vocabulary \cite{brown2008automated}.

\begin{table*}
	\centering 
	\footnotesize
	\begin{tabular}{c|ccccc}
		\toprule
		NER & De & En & Es & Fi & Nl \\ 
		\midrule
		Monolingual  & 79.78 $\pm$ .49 & 86.78 $\pm$ .15 & 78.99 $\pm$ .91 & 78.00 $\pm$ .87 & 78.91 $\pm$ .42 \\ 
		Multilingual & 75.37 $\pm$ .87 & 86.52 $\pm$ .34 & 78.33 $\pm$ .47 & 77.41 $\pm$ .86 & 77.49 $\pm$ .45 \\ 
		Mono + Multi & 81.13 $\pm$ .46 & \textbf{88.01 $\pm$ .27} & 80.32 $\pm$ .50 & 81.44 $\pm$ .36 & 81.15 $\pm$ .43 \\ 
		\midrule
		Mono + DE & - & 87.46 $\pm$ .19 & 79.79 $\pm$ .74 & 80.31 $\pm$ .21 & \bestBoth{81.31 $\pm$ .15} \\ 
		Mono + EN & 80.92 $\pm$ .29 & - & 80.48 $\pm$ .56 & \bestBoth{81.22 $\pm$ .26} & \bestPred{80.84 $\pm$ .30} \\ 
		Mono + ES & 80.29 $\pm$ .20 & 87.37 $\pm$ .30 & - & 80.80 $\pm$ .83 & 80.62 $\pm$ .39 \\ 
		Mono + FI & 81.10 $\pm$ .36 & \bestReal{87.94 $\pm$ .17} & 79.91 $\pm$ .82 & - & 80.65 $\pm$ .48 \\ 
		Mono + NL & \bestBoth{81.25 $\pm$ .14} & \bestPred{87.38 $\pm$ .22} & \bestBoth{\textbf{80.93 $\pm$ .25}} & 80.67 $\pm$ .49 & - \\ 
		\midrule
		Mono + All & 81.52 $\pm$ .33 & 87.70 $\pm$ .06 & 80.63 $\pm$ .34 & 82.07 $\pm$ .33 \ssBase & 81.73 $\pm$ .26 \ssBase \\ 
		Meta-Embeddings & \textbf{81.75 $\pm$ .50} \ssBase & 87.87 $\pm$ .23 & 80.84 $\pm$ .52 & \textbf{83.12 $\pm$ .12} \ssBase & \textbf{82.13 $\pm$ .50} \ssBase \\ 
		\bottomrule \\
		
	POS & De & En & Es & Fi & Nl \\ 
	\midrule
	Monolingual  & 93.02 $\pm$ .11 & 94.17 $\pm$ .09 & 96.23 $\pm$ .04 & 92.84 $\pm$ .13 & 94.01 $\pm$ .21 \\ 
	Multilingual & 92.19 $\pm$ .20 & 94.10 $\pm$ .06 & 96.01 $\pm$ .07 & 91.95 $\pm$ .11 & 93.35 $\pm$ .22 \\ 
	Mono + Multi & 93.40 $\pm$ .08 & 95.11 $\pm$ .07 & \textbf{96.54 $\pm$ .03} & 94.70 $\pm$ .12 & 94.94 $\pm$ .13 \\ 
	\midrule
	Mono + DE & - & \bestReal{95.11 $\pm$ .09} & 96.43 $\pm$ .13 & 94.43 $\pm$ .18 & \bestPred{94.70 $\pm$ .09} \\ 
	Mono + EN & 93.26 $\pm$ .11 & - & \bestReal{96.52 $\pm$ .06} & \bestPred{94.45 $\pm$ .14} & \bestPred{94.80 $\pm$ .12} \\ 
	Mono + ES & 93.31 $\pm$ .13 & 95.03 $\pm$ .09 & - & \bestReal{94.48 $\pm$ .14} & 94.79 $\pm$ .17 \\ 
	Mono + FI & 93.41 $\pm$ .12 & 94.97 $\pm$ .04 & 96.34 $\pm$ .08 & - & \bestReal{94.92 $\pm$ .13} \\ 
	Mono + NL & \bestBoth{93.52 $\pm$ .10} & \bestPred{94.99 $\pm$ .08} & \bestPred{96.41 $\pm$ .07} & 94.42 $\pm$ .08 & - \\ 
	\midrule
	Mono + All & \textbf{93.60 $\pm$ .14} \ssBase & \textbf{95.40 $\pm$ .04} \ssBase & 96.46 $\pm$ .09 & \textbf{95.61 $\pm$ .08} \ssBase & 95.31 $\pm$ .08 \\ 
	Meta-Embeddings & 93.51 $\pm$ .08 & 95.36 $\pm$ .10 \ssBase & 96.48 $\pm$ .06 & \textbf{95.61 $\pm$ .11} \ssBase & \textbf{95.34 $\pm$ .14} \ssBase \\ 
	\bottomrule
	\end{tabular}
	\caption{Results of NER ($F_1$, above) and POS (Accuracy, below) experiments with BPE embeddings. 
		\ssBase marks models that are statistically significant to the best Mono + X model. 
		\bestPred{box} highlights the closest auxiliary language according to language distance measure \dVoting, 
		and \bestReal{box} the best auxiliary language according to performance.
	}
	\label{tab:ner_bpe_results}
\end{table*}

\begin{table}
	\centering 
	\footnotesize
	\begin{tabular}{p{0.005cm}p{2.78cm}|p{0.4cm}p{0.4cm}p{0.4cm}p{0.4cm}p{0.4cm}}
		\toprule
		& & De & En & Es & Fi & Nl \\ 
		\midrule
		\multirow{3}{*}{\rotatebox{90}{NER}}&\newcite{ner/Strakova19} & \textbf{85.1} & \textbf{93.3} & \textbf{88.8} & - & \textbf{92.7} \\
		&Meta-Emb. (BPEmb)  & 81.8 & 87.9 & 80.8 & 83.1 & 82.1 \\ 
		&Meta-Emb. (FLAIR)  & 83.9 & 90.7 & 86.2 & \textbf{85.1} & 86.6 \\ 
		\midrule
		\multirow{3}{*}{\rotatebox{90}{POS}}&\newcite{yasunaga-etal-2018-robust}  & 94.4 & 95.8 & 96.8 & 95.4 & 93.1 \\
		&Meta-Emb. (BPEmb)  & 93.5 & 95.4 & 96.5 & 95.6 & 95.3 \\ 
		&Meta-Emb. (FLAIR)  & \textbf{94.8} & \textbf{96.5} & \textbf{97.2} & \textbf{97.8} & \textbf{96.8} \\ 
		\bottomrule
	\end{tabular}
	\caption{Comparison with state of the art.}
	\label{tab:sota_results}
\end{table}

\section{Sequence Tagging Model}\label{sec:approach}
Following \newcite{ner/Lample16}, we use a
bidirectional 
long short-term memory network (BiLSTM)
~\cite{lstm/Hochreiter97} followed by a
conditional random field (CRF) classifier \cite{lafferty01-crf} (see Figure~\ref{fig:model_base}).

\subsection{Embeddings}
Each input word is represented with a pretrained word vector.
We experiment with \bpeEmb (BPEmb)~\cite{bpemb/heinzerling18}
and \flairEmb embeddings~\cite{flair/Akbik18},
as for both of them pretrained embeddings are publicly available for all the languages we consider.\footnote{\url{https://github.com/flairNLP/flair} \url{https://nlp.h-its.org/bpemb/}}

\subsection{Combination of Embeddings} \label{sec:meta-concat}
As we experiment with multiple word embeddings, we compare two combination methods: 
a simple \textbf{concatenation} $e^{CONCAT}$ and attention-based \textbf{\metaEmb} $e^{ATT}$ as shown in Figure~\ref{fig:model_concat} and \ref{fig:model_meta}, respectively, and described next.

In both cases, the input are $n$ embeddings $e_i, 1 \le i \le n$ that should be combined. In our experiments, we use embeddings from $n$ different languages.

For concatenation, we simply stack the individual embeddings into a single vector: $e^{CONCAT} = [e_1, e_2, .., e_n]$.

In the case of \metaEmb, we follow~\newcite{kiela-etal-2018-dynamic} and compute the combination as a weighted sum of embeddings.
For this, all $n$ embeddings $e_i$ need to be mapped to the same size first. 
In contrast to previous work, we use a nonlinear mapping as this yielded better performance in our experiments.
Thus, we compute $x_i = tanh(Q_i \cdot e_i + b_i)$ with weight matrix $Q_i$, bias $b_i$ and $x_i \in \mathbb{R}^E$ being the $i$-th embedding $e_i$ mapped to the size $E$ of the largest embedding.
The attention weight $\alpha_i$ for each embedding $x_i$ is then computed with a fully-connected hidden layer of size $H$ with parameters 
$W \in \mathbb{R}^{H \times E}$ and
$V \in \mathbb{R}^{1 \times H}$, followed by a softmax layer.

\begin{equation*}
\alpha_i = \frac{\exp(V \cdot \tanh(W x_i))}{\sum_{l=1}^n \exp(V \cdot \tanh(W x_l))}
\end{equation*}

The parameters of the meta-embedding layer ($Q_1, ..., Q_n, b_1, ..., b_n, W, V$) are randomly initialized and learnt during training. 

Finally, the embeddings $x_i$ are weighted using the attention weights $\alpha_i$ resulting in the word representation $e^{ATT} = \sum_i \alpha_i \cdot x_i$.

\section{Experiments and Results}\label{sec:experiments}
We perform NER and POS experiments on five languages: German (De), English (En), Spanish (Es), Finnish (Fi), and Dutch (Nl). 
Note that we assume at least a character overlap to use auxiliary embeddings from another language.
Thus, languages with a different character set, e.g., Asian languages, cannot be used, in this setting.
Future work could investigate the inclusion of languages with different character sets, e.g., by using bilingual dictionaries or machine translation.

For NER, we use the CoNLL 2002/03 datasets \cite{data/conll/Sang02, data/conll/Sang03}
and the FiNER corpus~\cite{finer/Ruokolainen19}. 
For POS tagging, we experiment with the universal dependencies treebanks.\footnote{We predict the UPOS tag from the following UD v2.0 treebanks: de\_gsd, en\_ewt, es\_gsd, fi\_tdt, nl\_alpino.}
For each language,
we report results for the following methods:  

\paragraph{Monolingual (Mono).} 
Only embeddings from the source language were taken for the experiments. This is the baseline setting. 
We also compare our results to \textbf{multilingual embeddings (Multi)}
which have been successfully used in monolingual settings as well
~\cite{heinzerling-strube-2019-sequence}.
To ensure comparability, we use the multilingual versions of BPEmb and \flairEmb,
which were trained simultaneously on 275 and 300 languages, respectively.

\paragraph{Mono + X.}
A second set of embeddings from a different language X is concatenated with the original monolingual embeddings. 
While for this typically embeddings from a related language are chosen, we report results for all language combinations and investigate in particular whether relatedness is necessary for improvement. 

\paragraph{Mono + All \& Meta-Embeddings.}
We also experiment with the combination of all embeddings from all languages from our experiments. In this setting, we use all six embeddings (five monolingual embeddings and the multilingual embeddings) and combine them either using concatenation (Mono + All) or \metaEmb.
We have chosen these settings that are mainly based on monolingual embeddings, as the current state-of-the-art for named entity recognition is based on monolingual \flairEmb embeddings~\cite{flair/Akbik19}. 
In addition, multilingual embeddings, such as multilingual BERT~\cite{bert/Devlin19} tend to perform worse than their monolingual counterparts\footnote{\url{https://github.com/google-research/bert/blob/master/multilingual.md}} in monolingual experiments. For completeness, we include experiments with multilingual embeddings as mentioned before. 

\subsection{Results}\label{sec:results}
Following \newcite{reporting/ReimersGurevych17}, we report all experimental results as the mean of five runs
and their standard deviation in Table~\ref{tab:ner_bpe_results} for experiments with \bpeEmb embeddings.
The results with \flairEmb embeddings can be found in the appendix.
We performed statistical significance testing to check if the concatenation (Mono + All) and \metaEmb models are better than 
the best Mono + X model. 
We used paired permutation testing with 2$^{20}$ permutations and a significance level of 0.05 and performed the Fischer correction following \newcite{dror-etal-2017-replicability}.\footnote{We 
take the model with median performance on the development set for significance testing. }

For \metaEmb, we found statistically significant differences in 12 out of 20 settings (6 with BPEmb, 6 with FLAIR) against the best monolingual + X model, while we found statistically significant differences for Mono + All in only 7 out of 20 cases.
This suggests that \metaEmb are superior to monolingual settings with one auxiliary language as well as to the concatenation of all embeddings. 
Further, we found that the combination of monolingual and multilingual \bpeEmb embeddings is always superior to either monolingual or multilingual embeddings alone for both tasks. Even though the multilingual embeddings have seen many languages during pre-training, they can still benefit from the high performance of monolingual embeddings and vice versa. 
As the \metaEmb assign attention weights for each embedding, we can inspect the importance the models give to the different embeddings.
An analysis for an example sentence can be found in Section~\ref{app:att_weights} in the appendix.
Table \ref{tab:sota_results} provides the results of BPEmb and FLAIR meta-embeddings in comparison to state of the art, showing that we set the new state of the art for POS tagging.

\subsection{Analysis of Language Distances}

\begin{table}
	\centering 
	\footnotesize
	\begin{tabular}{c|ccccc}
		\toprule    
		\multicolumn{1}{c|}{\multirow{2}{*}{Rank}}  
		& \multicolumn{5}{c}{\dVoting}
		\\
		& De & En & Es & Fi & Nl 
		\\
		\midrule
		1 & Nl & Nl & En & En & De\fs
		\\
		2 & En & Fi & Nl & De & En\fs
		\\
		3 & Fi & De & Fi & Nl & Fi
		\\
		4 & Es & Es & Es & Es & Es
		\\
		\bottomrule
	\end{tabular}
	\caption{Language ranking according to the majority voting distance \dVoting. 
		\fs highlights equal ranks. }
	\setlength\tabcolsep{6pt}
	\label{tab:att_language_distances}
\end{table}

To evaluate how useful language distances are for predicting the best auxiliary language, we compare rankings based on language distances and the observed performance rankings based on Table~\ref{tab:ner_bpe_results}.
For this, we take the language distance from \newcite{language/Gamallo17}, which is based on language modeling perplexity $\text{PP}$ of unigram language models $\text{LM}$ applied to texts of foreign languages $\text{CH}$. Lower language model perplexities on a foreign dataset indicate higher language relatedness. 
\begin{equation*}
d_{P} (L1, L2) = \text{PP}(\text{CH}_{L2}, \text{LM}_{L1})
\end{equation*}

We also test language similarities based on vocabulary overlap 
with $W(L1|L2)$ being the number of words of $L1$ which are shared with $L2$ and $N(L1)$ the number of words of $L1$ shared with other languages.
\begin{equation*}
d_{V} (L1, L2) = \frac{W(L1|L2) + W(L2|L1)}{2 \cdot \min(N(L1), N(L2))}
\end{equation*}

\begin{figure}
	\centering
	\resizebox{.98\linewidth}{!}{\input{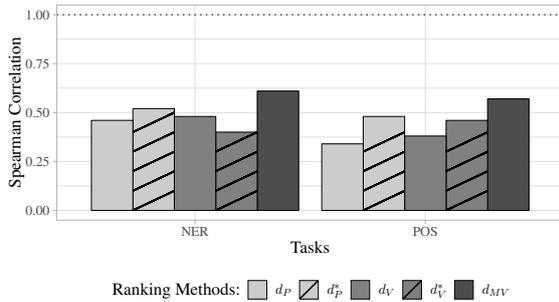}}
	\caption{Spearman's rank correlation between
		language distance and model performance rankings for NER and POS tasks
		for different language distances.}
	\label{fig:language_distance}
\end{figure}

For our experiments, we further adapt \dPerp to use the perplexity of the \flairEmb forward language models on the test data provided by \newcite{language/Gamallo17} and call it \dPerpFlair. 
Similarly, we adapt \dVocabTrain to compute the overlap of words in our training data.
Note that both variants, \dPerpFlair and \dVocabTrain, are based on properties from either our model or training data and are, therefore, specific to our setting. 
Finally, we create a ranking \dVoting which combines the rankings 
from \dPerp, \dPerpFlair, \dVocab, \dVocabTrain
with majority voting.
The ranking of \dVoting is provided in Table~\ref{tab:att_language_distances}, the rankings of the individual distance measures are given in the appendix. 

To analyze the correlation between language distance measures and the performance of our model, we compute the Spearman's rank correlation coefficient between the real rankings based on performance and predicted rankings from our language distances. The results are shown in Figure~\ref{fig:language_distance}. 
We conclude that predicting the auxiliary language ranking is a hard task
and see that the most related language is not always the best auxiliary language in practice (cf., Table~\ref{tab:ner_bpe_results}).
This holds in particular for POS tagging, where the performance differences of models are quite small.

In general, \dPerpFlair shows a higher correlation with model performance than \dPerp and \dVocab, indicating that not only word overlap plays a role but also context information.
The majority voting \dVoting achieves the highest correlation and often predicts the best auxiliary language for NER models using \bpeEmb. 
However, the actual ranking of all languages does not match the performance ranking, which results in a relatively low correlation with only a little above 0.5.

\subsection{Practical Guide}

\begin{figure}[h]
	\centering
	\includegraphics[width=.49\textwidth]{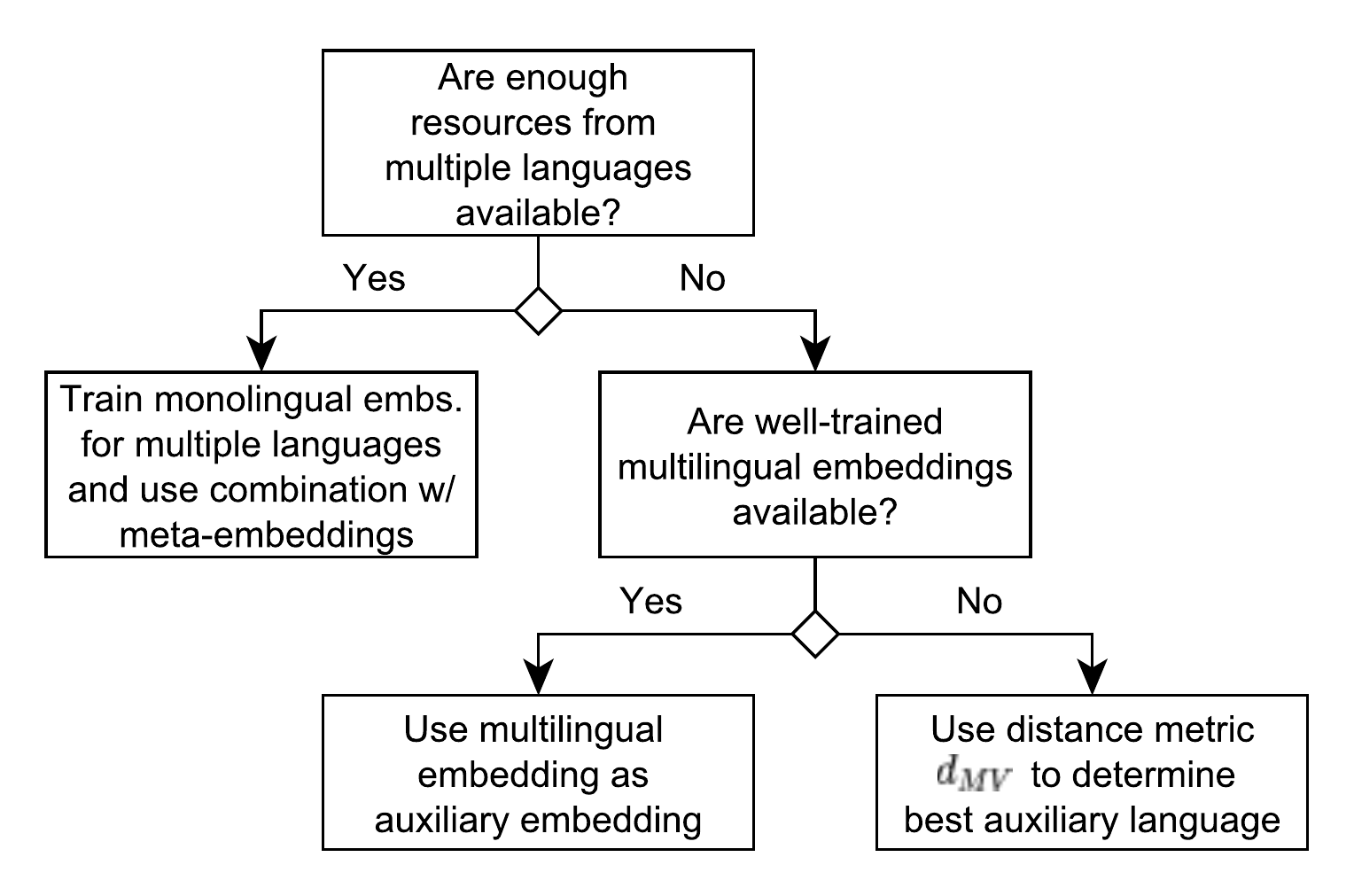}
	\caption{Proposal for auxiliary embedding selection. }
	\label{fig:decision}
\end{figure}

Finally, we propose a small guide in Figure~\ref{fig:decision} to decide which auxiliary languages one can use to improve performance over monolingual embeddings.
Depending on the available amount of data, it is recommended to train multiple monolingual embeddings and combine them using \metaEmb, which was observed to be the best method in our experiments. 
If not enough data is available to train monolingual embeddings, the best solution would be the inclusion of multilingual embeddings, assuming the existence of high-quality embeddings, such as multilingual \bpeEmb.
If none of the above applies, language distance measures, in particular the combination of multiple distances, can help to identify the most promising auxiliary embeddings.
Despite not always predicting the best model, the predicted auxiliary language often led to improvements over the monolingual baseline in our experiments.

\section{Conclusion}\label{sec:conclusion}
In this paper, we investigated the benefits of auxiliary languages for sequence tagging. We showed that it is hard to predict the best auxiliary language based on language distances.
We further showed that \metaEmb can leverage multiple embeddings effectively for those tasks and set the new state of the art on part-of-speech tagging across different languages. 
Finally, we proposed a guide on how to decide which method of including auxiliary languages one should use.

\section*{Acknowledgments}
We would like to thank Dietrich Klakow, Marius Mosbach, Michael A. Hedderich, the members of the BCAI NLP\&KRR research group and the anonymous reviewers for their helpful comments and suggestions.

\bibliography{refs}
\bibliographystyle{acl_natbib}

\appendix

\begin{table*}[hbt!]
	\setlength\tabcolsep{3.5pt}
	\centering 
	\footnotesize
	\begin{tabular}{c|cccc|cccc|cccc|cccc|cccc}
		\toprule		
		\multicolumn{1}{c|}{\multirow{2}{*}{Rank}}  & \multicolumn{4}{c|}{de} & \multicolumn{4}{c|}{en} & \multicolumn{4}{c|}{es} & \multicolumn{4}{c|}{fi} & \multicolumn{4}{c}{nl} \\
		\multicolumn{1}{c|}{} 
		& \dPerp & \dPerpFlair & \dVocab & \dVocabTrain
		& \dPerp & \dPerpFlair & \dVocab & \dVocabTrain
		& \dPerp & \dPerpFlair & \dVocab & \dVocabTrain
		& \dPerp & \dPerpFlair & \dVocab & \dVocabTrain
		& \dPerp & \dPerpFlair & \dVocab & \dVocabTrain \\
		\midrule
		1 
		& nl & nl & en & nl
		& nl & nl & de & fi
		& en & nl & en & en
		& de & nl & en & en
		& de & de & en & en \\
		2 
		& en & en & nl & en
		& es & fi & nl & nl
		& nl & en & de & nl
		& nl & de & de & nl
		& en & en & de & de \\
		3 
		& fi & fi & es\fs & fi
		& de & de & fi    & es
		& fi & de & fi    & fi
		& en & en & es\fs & de
		& fi & fi & es\fs & es \\
		4 
		& es & es & fi\fs & es
		& fi & es & es    & de
		& de & fi & nl    & de
		& es & es & nl\fs & es
		& es & es & fi\fs & fi \\
		\bottomrule
	\end{tabular}
	\caption{Language distances. Languages marked with \fs are ranked the same.}
	\setlength\tabcolsep{6pt}
	\label{tab:language_distances2}
\end{table*}

\section{Hyperparameters and Training}
We use the Byte-Pair-Encoding embeddings~\cite{bpemb/heinzerling18} with 300 dimensions and a vocabulary size of 200k tokens for all languages. For FLAIR, we use the embeddings provided by the FLAIR framework~\cite{flair/Akbik18} with 2048 dimensions for each language model resulting in a total embedding size of 4096 dimensions. 
The bidirectional LSTM has a hidden size of 256 units.
For training, we use stochastic gradient descent with a learning rate of 0.1 and a batch size of 64 sentences. The learning rate is halved after 3 consecutive epochs without improvement on the development set.
We apply dropout with probability 0.1 after the input layer.

\section{Language Distances}
We report the language rankings of the single metrics \dPerp, \dPerpFlair, \dVocab and \dVocabTrain in Table~\ref{tab:language_distances2}.

\section{Results on NER and POS tagging with FLAIR embeddings}
\label{sec:appendix:x}
We performed the same experiments as in Section 4.1 with FLAIR embeddings as well and report the results in Table~\ref{tab:pos_flair_results} for NER and for POS tagging.

In difference to the BPE experiments reported in the paper, we do not include multilingual embeddings in the Mono + All and meta-embedding versions of FLAIR. The reason is prior experiments in which multilingual embeddings led to reduced performance. This is also reflected in the poor performance of the multilingual FLAIR embeddings alone. It seems that the multilingual BPE embeddings are more effective in downstream tasks than the multilingual FLAIR embeddings. 

\begin{table*}[hbt!]
	\centering 
	\footnotesize
	\begin{tabular}{c|ccccc}
		\toprule
		NER & De & En & Es & Fi & Nl \\ 
		\midrule
		\newcite{ner/Strakova19} & 85.10 & 93.28 & 88.81 & - & 92.69 \\
		Monolingual & 82.66 $\pm$ .11 & 89.98 $\pm$ .11 & 85.08 $\pm$ .68 & 83.38 $\pm$ .31 & 85.68 $\pm$ .27 \\ 
		Multilingual & 66.21 $\pm$ .79 & 82.87 $\pm$ .24 & 77.87 $\pm$ .93 & 73.95 $\pm$ .74 & 77.44 $\pm$ .52 \\ 
		Mono + Mono & 82.45 $\pm$ .45 & 89.95 $\pm$ 0.21 & 85.26 $\pm$ .06 & 83.37 $\pm$ .48 & 85.67 $\pm$ .06 \\ 
		Mono + Multi & 82.95 $\pm$ .21 & 90.04 $\pm$ .11 & 84.70 $\pm$ .50 & 83.46 $\pm$ .37 & 86.04 $\pm$ .28 \\
		\midrule
		Mono + DE & - & \bestReal{90.24 $\pm$ .19} & 85.16 $\pm$ .23 & \bestReal{84.23 $\pm$ .22} & \bestPred{85.82 $\pm$ .22} \\ 
		Mono + EN & \bestReal{83.27 $\pm$ .36} & - & 85.53 $\pm$ .20 & \bestPred{84.10 $\pm$ .26} & \bestBoth{86.73 $\pm$ .09} \\ 
		Mono + ES & 82.85 $\pm$ .34 & 90.14 $\pm$ .13 & - & 83.88 $\pm$ .31 & 86.16 $\pm$ .09 \\ 
		Mono + FI & 83.10 $\pm$ .45 & 90.14 $\pm$ .09 & 85.06 $\pm$ .64 & - & 86.14 $\pm$ .31 \\ 
		Mono + NL & \bestPred{82.79 $\pm$ .24} & \bestPred{90.18 $\pm$ .15} & \bestBoth{85.77 $\pm$ .27} & 83.65 $\pm$ .31 & - \\ 
		\midrule
		Mono + All & 83.43 $\pm$ .29 & 90.29 $\pm$ .18 & 85.48 $\pm$ .78 & 84.32 $\pm$ .32 & 86.43 $\pm$ .33 \\
		Meta-Embeddings & 83.90 $\pm$ .14 \ssBase & 90.70 $\pm$ .29 \ssBase & 86.18 $\pm$ .35 & 85.09 $\pm$ .30 \ssBase & 86.58 $\pm$ .58 \\ 
		\bottomrule \\
		
		POS & De & En & Es & Fi & Nl \\ 
		\midrule
		\newcite{yasunaga-etal-2018-robust} & 94.35 & 95.82 & 96.84 & 95.40 & 93.09 \\
		Monolingual & 94.72 $\pm$ .07 & 96.28 $\pm$ .05 & 97.08 $\pm$ .03 & 97.52 $\pm$ .03 & 96.48 $\pm$ .11 \\ 
		Multilingual & 92.82 $\pm$ .20 & 93.69 $\pm$ .07 & 96.06 $\pm$ .13 & 92.98 $\pm$ .10 & 94.85 $\pm$ .11 \\ 
		Mono + Mono & 94.74 $\pm$ .15 & 96.24 $\pm$ .02 & 97.04 $\pm$ .08 & 97.55 $\pm$ .05 & 96.45 $\pm$ .13 \\ 
		Mono + Multi & 94.72 $\pm$ .13 & 96.29 $\pm$ .04 & 97.04 $\pm$ .05 & 97.52 $\pm$ .05 & 96.77 $\pm$ .02 \\ 
		\midrule
		Mono + DE & - & \bestReal{96.41 $\pm$ .07} & 97.11 $\pm$ .08 & \bestReal{97.64 $\pm$ .04} & \bestPred{96.62 $\pm$ .06} \\ 
		Mono + EN & \bestReal{94.71 $\pm$ .04} & - & 97.13 $\pm$ .12 & \bestPred{97.52 $\pm$ .06} & \bestPred{96.49 $\pm$ .09} \\ 
		Mono + ES & 94.67 $\pm$ .06 & 96.36 $\pm$ .03 & - & 97.48 $\pm$ .03 & 96.61 $\pm$ .13 \\ 
		Mono + FI & 94.65 $\pm$ .05 & 96.38 $\pm$ .03 & \bestReal{97.14 $\pm$ .05} & - & \bestReal{96.68 $\pm$ .05} \\ 
		Mono + NL & \bestPred{94.64 $\pm$ .03} & \bestPred{96.31 $\pm$ .07} & \bestPred{97.06 $\pm$ .04} & 97.51 $\pm$ .04 & - \\ 
		\midrule
		Mono + All & 94.64 $\pm$ .10 & 96.48 $\pm$ .05 & 97.11 $\pm$ .04 & 97.52 $\pm$ .06 & 96.54 $\pm$ .20 \\ 
		Meta-Embeddings & 94.78 $\pm$ .09 & 96.49 $\pm$ .03 \ssBase & 97.18 $\pm$ .07 & 97.82 $\pm$ .03 \ssBase & 96.83 $\pm$ .13 \ssBase \\ 
		\bottomrule
	\end{tabular}
	\caption{Results of NER ($F_1$, above) and POS (Accuracy, below) experiments with FLAIR embeddings. 
		\ssBase marks models that are statistically significant to the best Mono + X model. 
		\bestPred{box} highlights the closest auxiliary language according to language distance measure \dVoting,
		and \bestReal{box} the best auxiliary language according to performance.}
	\label{tab:pos_flair_results}
\end{table*}

\section{Analysis of Attention Weights}\label{app:att_weights}
\begin{figure}
	\centering
	\includegraphics[width=.48\textwidth]{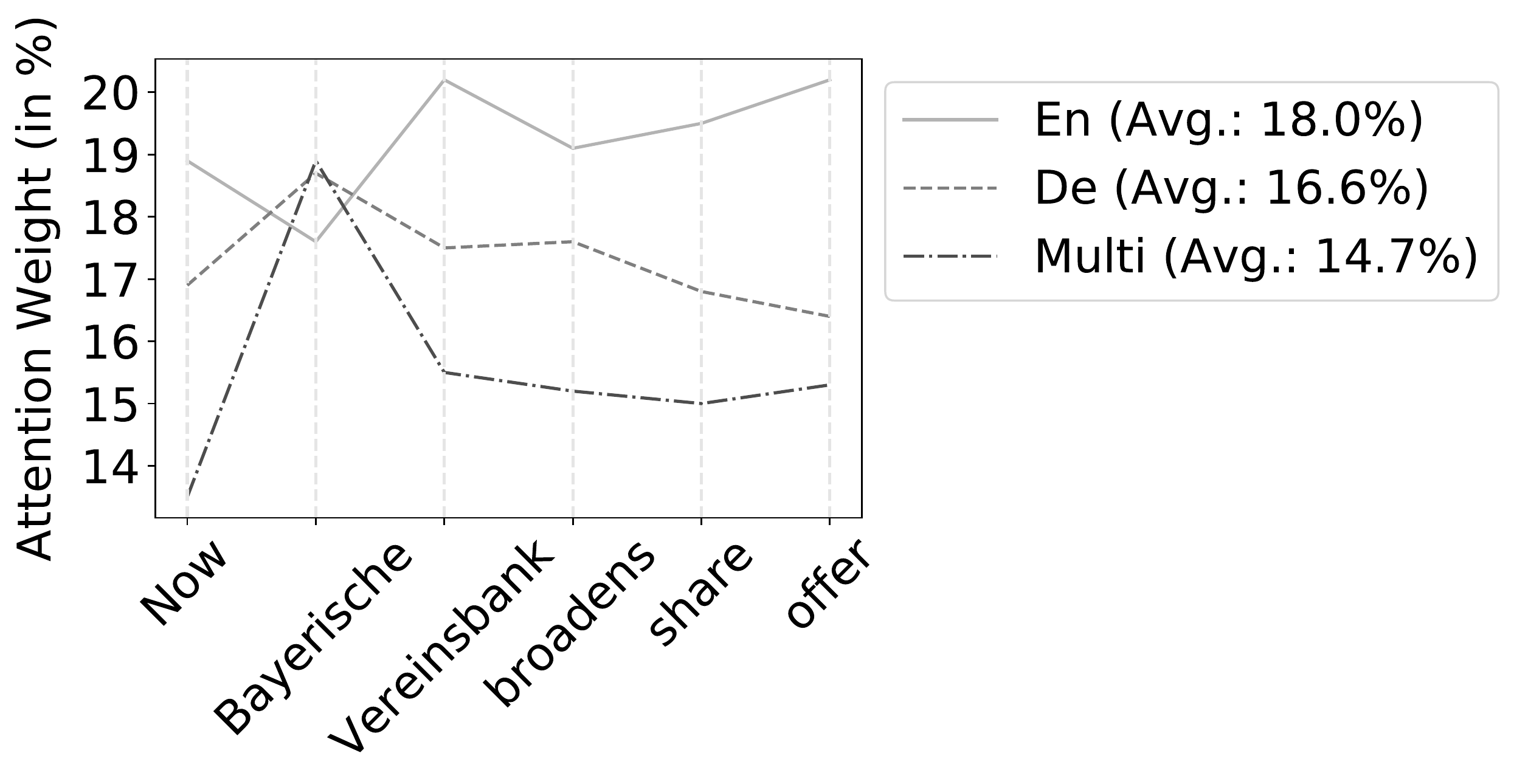}
	\caption{Learned attention weights of the \metaEmb model with \bpeEmb embeddings for English NER.}
	\label{fig:att_weights}
\end{figure}

As the \metaEmb assign attention weights for each embedding, we can inspect the importance the models give to the different embeddings. 
Figure~\ref{fig:att_weights} shows the assigned weights for an English sentence. In general, the model assigns most weight to the English embeddings. However, we observe an increased weight for German and the multilingual embedding for the German word \textit{Bayerische}. Even though \textit{Vereinsbank} is also a German word, the model focuses more on English for this word, probably because the subword \textit{bank} has the same meaning in English.

\section{Study: Increased Number of Parameters vs.\ Auxiliary Language}
To investigate whether the performance increase comes from the increased number of parameters rather than the inclusion of more embeddings, we also investigated including the same embedding type twice (Mono + Mono). However, we found that this does not help: The performance is comparable to the monolingual baseline. Thus, the performance increase for Mono + X really comes from additional information provided by the embeddings of the auxiliary language.

\end{document}